\documentclass{article}
\usepackage{iclr2027_conference,times}
\usepackage{float}
\iclrfinalcopy

\usepackage{hyperref}
\hypersetup{colorlinks=true, citecolor=blue, linkcolor=blue, urlcolor=blue}
\usepackage{url}
\usepackage{booktabs}
\usepackage{array}
\newcolumntype{L}[1]{>{\raggedright\arraybackslash}p{#1}}
\usepackage{amsmath}
\usepackage{amssymb}
\usepackage{graphicx}
\usepackage{microtype}
\usepackage{enumitem}
\setlist{nosep, leftmargin=1.4em}
\usepackage{tikz}
\usetikzlibrary{positioning,arrows.meta,fit,calc,shapes.misc}
\newcommand{\zd}{z_\Delta}
\newcommand{\dy}{|\Delta\hat{y}|}
\newcommand{\dexcess}{\Delta_{\mathrm{excess}}}

\title{Causal Structure is Inducible but Functionally Decoupled: The Routing/Readout Boundary of a Typed Mechanism Library}

\author{Xining Xun \\
Tsingjiao Information Science (Beijing) Co., Ltd.
}

\begin{document}

\maketitle
\lhead{Preprint}

\begin{abstract}
When a language model answers an interventional question, the computation it must perform depends on the \emph{type of evidence} the query requires --- different types are different computations. We report a \textbf{decoupling} in how a transformer organizes causal knowledge: slot$\times$type structure induced by type-level supervision organizes \emph{routing}, yet remains functionally decoupled from answer \emph{readout}. We establish this with a \textbf{typed mechanism library} --- discrete mechanism slots, partitioned by evidence type, addressable and auditable at the state level --- used as a measurement instrument on a causal-world benchmark with exact interventional ground truth, under a frozen evaluation protocol, at two scales (22.6M and 125M). Four preregistered findings. \textbf{(i) Origin.} Slot$\times$type organization is \emph{induced} by type-level supervision: it does not emerge in architecturally identical unsupervised controls, it cannot be bought by content-free gating labels, and where it appears it is statistically attributable to the supervision signal, replicating at 125M under a powered, preregistered attribution protocol (three fresh seeds, two fresh unsupervised controls, 1,280 fresh queries; gate passed on all nine criterion cells). \textbf{(ii) Boundary.} The induced structure is a \textbf{typed routing index} with a sharp \textbf{routing/readout boundary}: structural slot codes scaffold routing but do not drive answer readout ($\dy \le 3.4\times10^{-6}$ with exactly zero collateral, three seeds, stable across a 5.6$\times$ scale window) --- we therefore make no behavioral-editability claim. \textbf{(iii) Cost.} The structure is free: LM quality matches a parameter-matched monolith within 0.0082 nats (preregistered tolerance, three seeds; behavioral battery 50/50). \textbf{(iv) Trust.} The library state is exactly local under edit and \textbf{bit-exactly revertible} --- 250 single-edit and 1,000 stacked reverts per seed, three seeds, zero failures --- an auditable state-management substrate for causal knowledge, with properties that hold by construction and are verified under preregistered criteria. We further find that \textbf{the unsupervised null itself moves with scale} --- a moving null --- which scopes how organization claims may be inherited across scales (\S\ref{sec:organization}): architecture comparisons that reuse a null calibrated at one scale may be confounded at another. Every claim above is tied to a preregistered, machine-checkable criterion archived before the data it governs; the full audit trail --- including one criterion we failed and how the frozen protocol handled it --- is released as an appendix.
\end{abstract}

\section{Introduction}

A large part of causal-reasoning and interpretability research rests on an implicit assumption we call the \textbf{circuit-board intuition}: if we can expose a discrete internal structure --- a slot, a neuron, a circuit --- and modify it, the modification should show up in the model's output. Locate-and-edit methods write into localized weights; probing and circuit analyses read structure out and treat it as the computation. The intuition is load-bearing: without it, exposing structure buys understanding but no control. Is it true? We built a controlled platform to test the assumption end to end, and the answer is no --- in a precise, measured, and replicated sense.

The platform is a \textbf{causal-world} testbed with exact interventional ground truth, plus a \textbf{typed mechanism library} (MM) on a standard transformer backbone \citep{vaswani2017attention}: $N$ discrete slots, statically partitioned over evidence types (identity, relation, sign, \dots), with a gating head trained by a small auxiliary loss encouraging type-consistent routing. Real text corpora confound evidence type with topic, lexical cues, and frequency; here every world is a small structural causal model, interventional ground truth is computable exactly, and every query carries a known evidence type --- at two scales (22.6M and 125M parameters). One architectural choice is what makes the coming negative result informative rather than trivial: the library is the model's \textbf{only writable causal state}, and the backbone consumes library-mediated representations --- the architecture gives the explicit structure every condition to sit on the readout pathway. What training does with that opportunity is the empirical question.

Three questions organize the paper. \textbf{Where does typed organization come from} --- does it emerge for free, can it be bought by arbitrary supervision, or is it induced by supervision that is \emph{about} something? \textbf{What is the structure for} --- does it sit on the pathway that produces answers, or is it an epiphenomenal index? And \textbf{what does it cost and what can be trusted} --- does the structure degrade the base model, and can its state be edited and audited exactly? Our aim is \textbf{not} to improve causal reasoning per se --- every capability result in this paper is a parity result --- but to measure, under preregistered criteria, what explicit causal state can and cannot deliver.

\textbf{Contributions.} The findings come first; the instrument that produced them is the third contribution; the protocol is how we make all of it believable.

\begin{enumerate}
\item \textbf{Scientific finding: the routing/readout boundary (H-$\alpha$).} Induced slot$\times$type structure scaffolds routing but does not drive answer readout: 150/150 structural edits move the readout by $\le2\times10^{-4}$ with exactly zero collateral ($\dy \le 3.4\times10^{-6}$ on the criterion protocol, three seeds), stable across a 5.6$\times$ scale window. The boundary falsifies the circuit-board intuition for explicit-state architectures --- and it is not an architectural triviality: training decoupled a structure that was built to participate.
\item \textbf{Origin, attributed: induced, not emergent, not buyable.} Slot$\times$type organization is induced by type-level supervision (permutation-tested MI): content-free gating labels are unlearnable (two protocols, four seeds, 22.6M); unsupervised controls show no organization at 22.6M; and the supervised effect replicates across three seeds (z = 4.47/5.99/7.02, Stouffer z = 10.1; excess 0.072--0.124 nats, moderate effect grade per frozen tiering). At 125M the attributable increment replicates against two fresh unsupervised controls under a preregistered paired-attribution gate: three fresh type seeds clear the amplitude bar (z = 15.15/13.20/13.77; excess 0.1275/0.1025/0.1202 nats against the frozen $\ge 0.10$), and all six paired contrasts clear the frozen $\zd \ge 3$ bar (5.02--7.87; archive A20.21).
\item \textbf{Methodological caution: the moving null.} The unsupervised null is not scale-free: a structural-prior control sits on the null at 22.6M but carries weak, stable organization at 125M (z=2.38; \S\ref{sec:organization}). Organization claims calibrated at one scale and evaluated at another may inherit a null that no longer applies. We release the paired-permutation attribution protocol built to survive this --- frozen intersection gate, frozen edge clause prescribing exactly one remedy (a powered replication frozen before any replication data) --- as a reusable template.
\item \textbf{Instrument: a zero-cost, auditable state-management substrate.} The typed library (600 slots, 6 evidence types at 125M) costs nothing measurable in LM quality (gap $\le 0.0082$ nats at a preregistered tolerance; battery 50/50 both arms); edits have exactly zero off-path collateral (three seeds); and the revert audit passes in full --- single-edit and depth-20 stacked revert, backbone zero-touch, zero-contamination contract for failed edits, three seeds, zero failures.
\end{enumerate}

\section{Architecture}
\label{sec:architecture}

\begin{figure}[t]
\centering
\resizebox{\textwidth}{!}{%
\begin{tikzpicture}[
  font=\small,
  every node/.style={align=center},
  blk/.style={draw, rounded corners=2pt, thick, inner sep=5pt},
  gry/.style={blk, fill=black!6, draw=black!50},
  bb/.style={blk, fill=blue!7, draw=blue!60!black},
  gate/.style={blk, fill=yellow!20, draw=orange!80!black},
  ed/.style={blk, fill=green!8, draw=green!55!black},
  lib/.style={draw=blue!60!black, very thick, rounded corners=4pt, fill=blue!3,
              inner sep=7pt},
  tblock/.style={rounded corners=2pt, inner sep=4pt, font=\scriptsize\bfseries,
                 text=white, minimum width=1.55cm, minimum height=0.62cm},
  arr/.style={-{Stealth[length=2.6mm]}, very thick, draw=black!65},
  lbl/.style={font=\scriptsize, text=black!70},
]
\node[tblock, fill=blue!75]                           (t1) {identity};
\node[tblock, fill=green!65!black,  right=1.6mm of t1] (t2) {child};
\node[tblock, fill=yellow!75!orange, text=black, right=1.6mm of t2](t3) {relation};
\node[tblock, fill=orange!90!black,  right=1.6mm of t3](t4) {sign};
\node[tblock, fill=pink!65,          text=black, right=1.6mm of t4](t5) {confidence};
\node[tblock, fill=black!45,         right=1.6mm of t5](t6) {reserved};
\coordinate (rowmid) at ($(t3.north)!0.5!(t4.north)$);
\node[font=\footnotesize\bfseries, text=blue!60!black, anchor=south]
  (libtitle) at ([yshift=2.2mm]rowmid)
  {Mechanism library --- 600 typed slots
   \textnormal{\mdseries\scriptsize\;(the only writable causal state)}};
\pgfdeclarelayer{bg}
\pgfsetlayers{bg,main}
\begin{pgfonlayer}{bg}  
\node[lib, fit=(t1)(t6)(libtitle)] (lib) {};
\end{pgfonlayer}
\node[gate, above=10mm of lib] (gate)
  {Typed gating head\\
   \scriptsize $\lambda_g{=}0.1$ (type) $\cdot$ $\lambda_g{=}0$ (emergent / blocks)\\
   \scriptsize Gumbel $\tau$ anneal $\cdot$ $\beta$-floor 0.3 $\cdot$ $\lambda_{lb}$ 0.01};
\node[ed, below=10mm of lib] (edit)
  {\textbf{EditSession}\\
   \scriptsize snapshot $\to$ edit $\to$ undo\\
   \scriptsize flip\_sign $\cdot$ add\_edge $\cdot$ remove\_edge $\cdot$ swap\_edge $\cdot$ param\_edit};
\node[bb, left=11mm of lib] (bb)
  {Causal LM backbone\\ 116.66M params\\ \scriptsize (frozen after training)};
\node[gry, left=7mm of bb] (tok) {evidence\\ tokens};
\node[bb, right=20mm of lib] (head) {Answer\\ readout\\ head};
\draw[arr] (tok) -- (bb);
\draw[arr] (bb.north) |- node[pos=0.78, above, lbl] {routing logits} (gate.west);
\draw[arr] (gate.south) -- (lib.north);
\draw[arr] (lib.east) -- node[lbl, below=1pt] {library-mediated\\ features} (head.west);
\draw[arr, dashed, draw=orange!85!black]
  ([yshift=2.4mm]lib.east) --
  node[midway, cross out, draw=orange!85!black, very thick, minimum size=3mm,
       inner sep=0pt, fill=white] {}
  ([yshift=2.4mm]head.west);
\node[font=\scriptsize, text=orange!85!black, anchor=south]
  at ([yshift=2.5mm]head.north)
  {$\times$ structural codes NOT on readout path\\
   (H-$\alpha$, scale-invariant 22.6M $\to$ 125M)};
\draw[arr, draw=green!55!black] (edit.north) --
  node[lbl, text=green!40!black, right=3pt, align=left]
  {library-only writes (any slot) $\cdot$ backbone zero-touch\\
   bit-exact rollback: 250/250 edits + 1{,}000/1{,}000 levels $\times$3 seeds (A22)}
  (lib.south);

\end{tikzpicture}}
\caption{Architecture overview (mm125: 126.21M total; N=600 slots, 6 evidence types). Green path: edit interface with audited bit-exact rollback (A22, \S\ref{sec:revert}). Dashed orange path: the measured H-$\alpha$ boundary (\S\ref{sec:boundary}).}
\label{fig:architecture}
\end{figure}
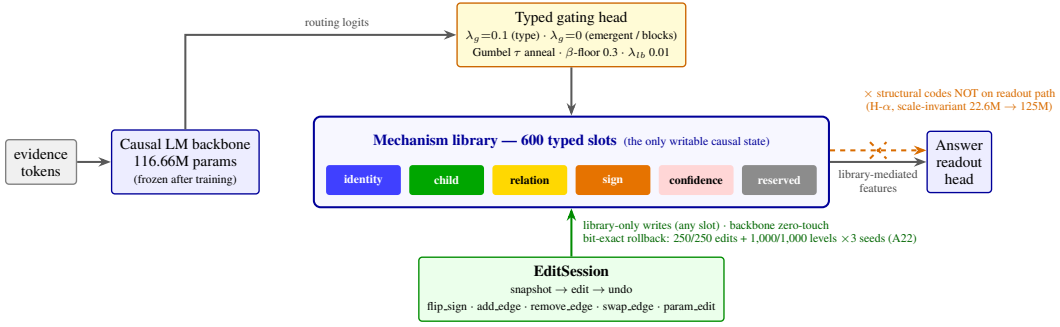

\subsection{Mechanism library}

The library is a tensor of N typed slots (N=600 at the 125M configuration, N=200 at 22.6M). Slots are statically partitioned over evidence types (6 types at 125M: identity / child / relation / sign / confidence / block-reserved; 3 types at 22.6M), with a per-type floor ($\beta_{\mathrm{floor}}=0.3$) preventing collapse. Each slot carries a mechanism payload (edge list, sign bits, scalar parameters) and audit buffers (usage counters) that are part of the model's \texttt{state\_dict} and therefore of every snapshot.

\subsection{Typed routing and gating}

Evidence tokens are routed to slots by a gating head. Three gate modes define our arm structure: \textbf{type} (supervised by the type-classification auxiliary loss $\lambda_g=0.1$), \textbf{emergent} ($\lambda_g=0$, free routing), and \textbf{blocks} ($\lambda_g=0$, block-structured routing prior without type labels). The two unsupervised arms are the null model for the organization claim: any slot$\times$type alignment they exhibit is organization that supervision cannot be credited for. Gumbel noise \citep{jang2017categorical,maddison2017concrete} ($\tau$ annealed over the first 1,500 steps) shapes early exploration; an entropy floor and load-balancing term ($\lambda_{lb}=0.01$) prevent degenerate routing.

\subsection{Edit interface}
\label{sec:editinterface}

Edits are applied through an \texttt{EditSession}: snapshot library state $\to$ apply operator $\to$ (optionally) verify $\to$ commit or undo. Four structural operators (\texttt{flip\_sign}, \texttt{add\_edge}, \texttt{remove\_edge}, \texttt{swap\_edge}) plus a bounded parameter operator (\texttt{param\_edit}, $\le$50 steps, row-masked lr $10^{-3}$) cover the library's writable dimensions. Composition is native: \texttt{compose\_legal} validates operator sequences against library invariants before application. All operators are human-readable state transformations --- but what they change behaviorally is bounded by H-$\alpha$ (\S\ref{sec:boundary}), and we scope our claims accordingly.

\subsection{Backbone and training}

A 116.66M-parameter causal-LM backbone (total 126.21M with library and heads at mm125) is trained on online-sampled causal worlds with the standard LM loss plus answer head, gating, and load-balancing terms. All arms share the frozen recipe: 40k steps, lr $6\times10^{-4}$ cosine to $6\times10^{-5}$ with 1.5k warmup, $\beta_{\mathrm{floor}}=0.3$, Gumbel 1.0, $\lambda_{lb}=0.01$; arms differ only in \texttt{gate\_mode} and $\lambda_g$. A parameter-matched monolithic transformer (MONO, 125M) serves as the LM-quality baseline.

\section{Experimental protocol}
\label{sec:protocol}

\textbf{Criteria first.} Every headline claim is paired with a preregistered, machine-checkable criterion archived (with md5-chained scripts) before the data it governs. Permutation tests (2,000 permutations, archived seeds; \citealp{pesarin2010permutation}) drive all mutual-information claims; editing claims are pass/fail with zero statistical degrees of freedom. A pipeline-validation gate accompanies every MI-based criterion: unsupervised control arms must sit within the null band ($|z|<2$) for the criterion to be interpretable --- a lesson burned into the protocol by an earlier incident in which a leaky pipeline \emph{fabricated} significance (Appendix~\ref{app:cases}, Case A12).

\textbf{When a criterion fails.} The protocol specifies in advance what happens next: suspension of the affected claim, a frozen diagnostic battery, and --- where warranted --- a revised estimand whose thresholds are frozen before the data it governs. This path was executed once in this project (the 125M pipeline-gate failure behind \S\ref{sec:organization}'s branching); the full decision tree as executed --- including the miss we did not re-judge --- is shown in Appendix~\ref{app:cases} (Figure~\ref{fig:tree}).

\begin{table}[t]
\caption{Preregistered criteria and current verdicts (full archive in Appendix~\ref{app:archive}).}
\label{tab:criteria}
\centering\small
\setlength{\tabcolsep}{4pt}
\begin{tabular}{@{}L{2.5cm}L{2.3cm}L{3.2cm}L{4.6cm}@{}}
\toprule
Criterion & Claim under test & Bar (frozen) & Verdict \\
\midrule
S1 (22.6M) & typed organization & per-seed $z\ge3$; mean-excess tiers frozen (strong: $\ge0.10$) & \textbf{PASS --- moderate grade} (z=4.47/5.99/7.02; excess 0.072--0.124, mean 0.0998) \\
S1 (125M raw) & typed organization & same, 3 seeds & raw PASS (z 5.62--7.75) --- attribution resolved by A20-R \\
A20-R (125M attribution replication) & attributable increment over unsupervised controls & per-seed $z\ge3 \wedge \mathrm{excess}\ge0.10$; 6/6 paired $\zd\ge3$ & \textbf{PASS} (z 13.20--15.15; excess 0.1025--0.1275; $\zd$ 5.02--7.87; archive A20.21) \\
Pipeline gate & MI pipeline integrity & unsupervised arms $|z|<2$ & blocks z=2.38 $\to$ attribution protocol (\S\ref{sec:attribution}) \\
S2 & pathway boundary / locality & $\dy\le10^{-3} \wedge \mathrm{collateral}<0.01$ & \textbf{PASS} ($\le3.4\times10^{-6}$; 0.0 exactly) \\
S3 & zero LM cost & $\mathrm{gap}\le0.02 \wedge \mathrm{battery}\ge95\%$ & \textbf{PASS} ($\le0.0082$; 50/50) \\
A22 & bit-exact rollback & R1--R5 all pass, $\times3$ seeds & \textbf{PASS} \\
A21 & persistence under continued training & C1--C3 (frozen) & [PENDING: unfrozen 2026-08-06 (A20.12 condition met); execution to be scheduled] \\
\bottomrule
\end{tabular}
\end{table}

\begin{table}[t]
\caption{Arms and frozen recipe (125M unless noted).}
\label{tab:arms}
\centering\small
\setlength{\tabcolsep}{4pt}
\begin{tabular}{@{}L{1.5cm}L{1.9cm}L{0.8cm}L{3.2cm}L{4.4cm}@{}}
\toprule
Arm & \texttt{gate\_mode} & $\lambda_g$ & role & seeds \\
\midrule
type & \texttt{type} & 0.1 & treatment & s0--s2 (archived) + s3--s5 (A20-R confirmatory) \\
emergent & \texttt{emergent} & 0 & unsupervised control & s0 (archived) + s1 (A20-R) \\
blocks & \texttt{blocks} & 0 & structural-prior control & s0 (archived) + s1 (A20-R control) + s2 (A20-R robustness, non-criterion) \\
MONO & --- & --- & LM-quality baseline (125M monolith) & s0 \\
\bottomrule
\end{tabular}
\end{table}

All arms: 40k steps, lr $6\times10^{-4}$ cosine $\to 6\times10^{-5}$ (warmup 1.5k), $\beta_{\mathrm{floor}}=0.3$, Gumbel 1.0, $\tau_{\mathrm{warmup}}=1500$, $\lambda_{lb}=0.01$; mm125 = 116.66M backbone + library (126.21M total), N=600, 6 types; mm25 = 22.6M, N=200, 3 types.

\textbf{Arms and seeds.} At 125M: type $\times6$ seeds (s0--s2 archived; s3--s5 A20-R confirmatory), emergent $\times2$ (s0 archived; s1 A20-R control), blocks $\times3$ (s0 archived; s1 A20-R control; s2 A20-R robustness-disclosure member --- not a criterion cell), MONO $\times1$ (LM baseline). At 22.6M: the same arm family at N=200 / 3 types.

\textbf{Anti-rescue discipline.} No re-judging archived values, no threshold edits after data, no seed top-ups, no second replications. Every protocol revision is archived \emph{before} the data it affects. The single revision of this project (binary pipeline check $\to$ paired attribution gate) followed this rule, and the event is reported in full (\S\ref{sec:attribution}, Appendix~\ref{app:cases}).

\section{Results}

\subsection{The moving null, and the origin of typed organization (criterion S1 + attribution gate)}
\label{sec:organization}

\textbf{22.6M (archived, final).} Type-supervised routing produces slot$\times$type organization that replicates across three seeds: per-seed z = 4.47 / 5.99 / 7.02 (each $p\le5\times10^{-4}$; Stouffer z = 10.1; \citealp{stouffer1949american}), debiased MI excess 0.072--0.124 nats (mean 0.0998), while both unsupervised arms sit on the null (emergent z=+0.21, blocks z=+0.14; T3 archive pull). Per the frozen tiering, the verdict is \textbf{PASS at the moderate grade}: the per-seed z gate was met 3/3; the strong-grade mean-excess bar (0.10) was missed by 0.0002 and is recorded as such (\S\ref{sec:discussion}, Limitations). A control condition with semantically arbitrary labels was not learnable at all (archived control F1, two protocols, four seeds): the gating head exploits genuine type structure, not label noise.

\textbf{125M: the null moved.}
\label{sec:baselines}
All three type seeds pass the raw amplitude gate: z = 6.09 / 7.75 / 5.62, excess = +0.141 / +0.164 / +0.134 nats (seed-aligned; 320 queries, 2,000 permutations). The pipeline-validation gate, however, did \textbf{not} pass cleanly: the blocks arm measured z=+2.38 (excess +0.050, p=0.012), above the $|z|<2$ null band, while emergent sat at z=+1.09. Per the preregistered decision tree, S1 at 125M was suspended pending diagnosis. The diagnostic finding is itself a result of this paper: \textbf{the unsupervised null is not scale-free.} At 22.6M both unsupervised arms sit exactly on the null (emergent z=+0.21, blocks z=+0.14); at 125M the blocks arm carries a small, stable slot$\times$type alignment (+0.050 nats; z=2.38, cross-permutation-seed sd 0.031). A fresh blocks seed settles the level at which the effect lives (D-d, blocks s1: z=+3.89, excess +0.065 nats, criterion protocol): the elevation replicates across seeds of the same arm --- it is \textbf{arm-level}, i.e., the blocks routing prior itself carries weak slot$\times$type organization at 125M (s0: +0.050; s1: +0.065), not a seed-level fluctuation. Because scale and type-count changed together (N 200$\to$600, 3$\to$6 types), the baseline shift's cause is \textbf{confounded in our data and we report it as such} --- we do not decompose what we cannot decompose. Either way, the methodological lesson stands: an organization claim at one scale cannot be silently inherited by another scale, because the null itself can move. We also record, as a \textbf{post-hoc observation} (archived as such; not a preregistered hypothesis), that all three 125M excess values exceed the 22.6M range --- an apparent $\approx$1.5$\times$ amplification of effect size with scale whose cause we do not decompose.

\begin{figure}[t]
\centering
\includegraphics[width=\textwidth]{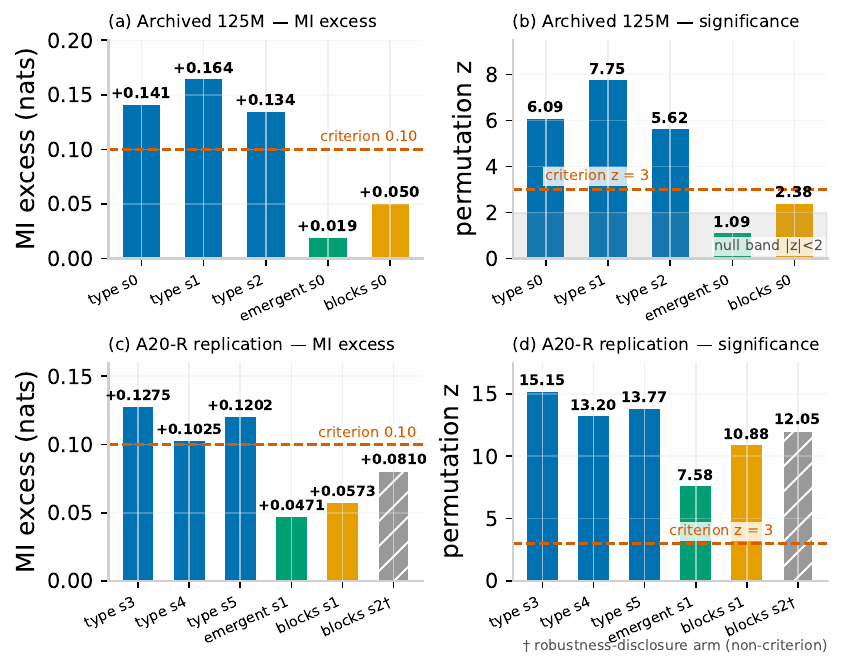}
\caption{Organization at 125M, archived (a,b) and replication (c,d). Type seeds clear both bars in both collections; the blocks arm's small but nonzero elevation (archived z=2.38) is what triggered the attribution protocol below, and the replication controls sit higher still --- the moving null in raw form. The hatched blocks s2 arm is the robustness-disclosure member and does not enter the gate.}
\label{fig:organization}
\end{figure}

\begin{figure}[t]
\centering
\includegraphics[width=\textwidth]{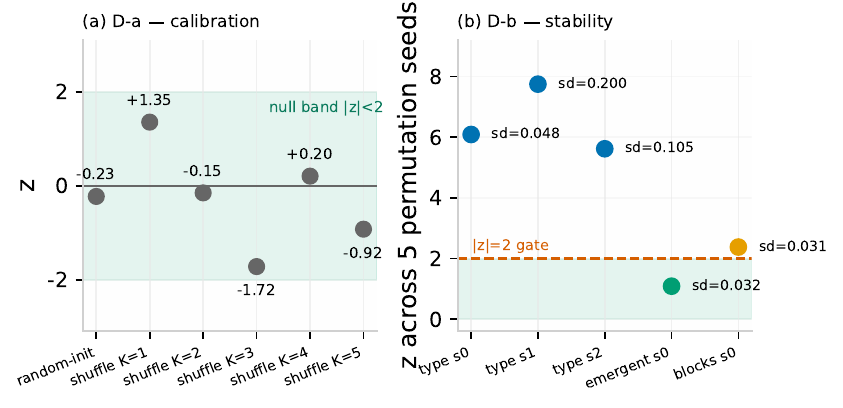}
\caption{The pipeline is calibrated and the offending measurement is stable --- the basis on which the protocol concluded ``the null moved'' rather than ``the pipeline broke''. (a) D-a calibration: random-init model and five label-shuffled audits all inside the null band. (b) D-b stability: z across five permutation seeds per arm; the blocks elevation is tight (sd 0.03), i.e., a real measurement, not a permutation artifact.}
\label{fig:diagnostics}
\end{figure}

\textbf{Attribution: revising the estimand, not the bar.}
\label{sec:attribution}
Three preregistered checks established the diagnosis: (i) the pipeline is calibrated --- a random-init model gives z=$-0.227$ and label-shuffled audits give $z\in[-1.72,+1.36]$, all within null; (ii) the blocks elevation is a stable measurement, not a permutation artifact --- across the archived permutation seed plus four fresh ones (five values per arm), blocks $z\in[2.34,2.42]$ (sd 0.031); (iii) a paired attribution test shows the type-supervised increment over each control is positive on all six seed$\times$control contrasts ($\dexcess$ +0.085\dots+0.145 nats). The binary pipeline check (unsupervised $|z|<2$) assumes the unsupervised null is zero. When that assumption failed, we did not lower any bar; we revised the \emph{estimand}. The \textbf{paired-permutation attribution test} applies the \emph{same} label permutation to both arms synchronously and normalizes the arm difference: $\zd = (\Delta - \mathrm{null\_mean})/\mathrm{null\_sd}$ over 2,000 paired permutations. On archived data (320 queries), five of six contrasts clear $\zd\ge3$, and the weakest (s2 vs blocks) reaches 2.68 (p=0.0035). Because the protocol froze an intersection rule, the archived verdict is ``gate not met''; because the effect sizes are uniformly positive and the miss is marginal, the protocol's own edge clause opened exactly one remedy --- a powered replication on fresh seeds with fresh worlds (A20-R: n\_worlds=160 $\to$ 1,280 queries, gate unchanged, protocol frozen before any replication data). We release the test, its gates, and its edge clause as a reusable template for organization claims whose unsupervised null may be nonzero.

\begin{table}[t]
\caption{Paired attribution contrasts (archived, 125M). $\dexcess$ = type-arm MI excess $-$ control-arm MI excess.}
\label{tab:attribution}
\centering\small
\setlength{\tabcolsep}{6pt}
\begin{tabular}{@{}lccl@{}}
\toprule
Contrast & $\dexcess$ (nats) & $\zd$ & $p$ \\
\midrule
s0 vs blocks & +0.091 & +3.15 & $<$1e-4 \\
s0 vs emergent & +0.122 & +4.47 & $<$1e-4 \\
s1 vs blocks & +0.114 & +3.89 & $<$1e-4 \\
s1 vs emergent & +0.145 & +5.21 & $<$1e-4 \\
s2 vs blocks & +0.085 & \textbf{+2.68} & 0.0035 \\
s2 vs emergent & +0.116 & +3.92 & $<$1e-4 \\
\bottomrule
\end{tabular}
\end{table}

\begin{figure}[t]
\centering
\includegraphics[width=\textwidth]{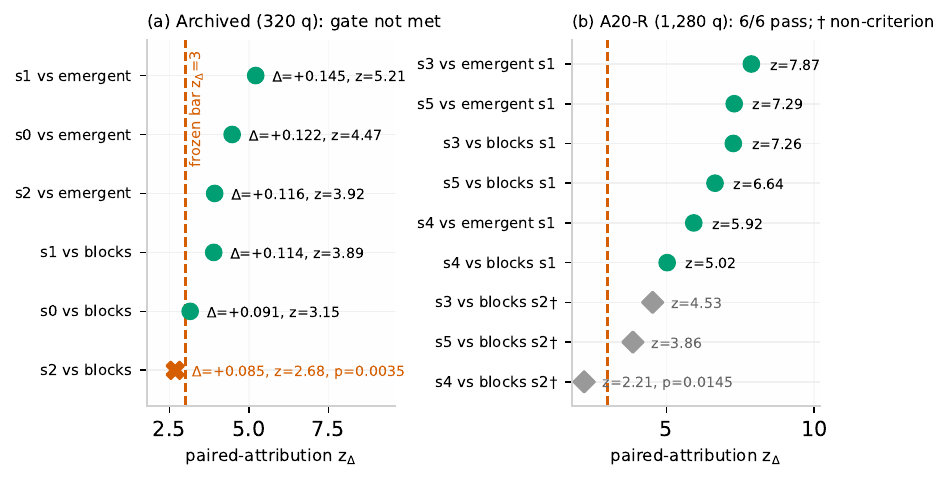}
\caption{Attribution contrasts, archived (a) and replication (b), with the frozen $\zd=3$ bar. (a) Archived collection (320 queries): five of six contrasts pass, one (weakest seed $\times$ strongest baseline) lands at 2.68 and the intersection gate is not met. (b) A20-R replication (1,280 queries): all six criterion contrasts pass; gray diamonds are the robustness-disclosure contrasts against blocks s2 (non-criterion). The intersection rule makes no exception for the 2.68 --- and neither do we; the remedy is the powered replication, not a lower bar.}
\label{fig:attribution}
\end{figure}

\textbf{Replication (A20-R, verdict 2026-08-06; archive A20.21).} The powered replication (three fresh type seeds, two fresh unsupervised controls, 1,280 fresh queries, fresh world set; gate: per-seed amplitude $z\ge3 \wedge \mathrm{excess}\ge0.10$, and paired $\zd\ge3$ against \emph{each} control) executed exactly as frozen, and the gate passed on all nine criterion cells (Table~\ref{tab:a20r}) --- this section's branch was selected mechanically, not argued.

\begin{table}[t]
\caption{A20-R confirmatory replication (125M; n\_worlds=160 $\to$ 1,280 fresh queries; 2,000 paired permutations; fresh world set disjoint from the archived criterion set). All nine criterion cells pass the frozen intersection gate.}
\label{tab:a20r}
\centering\small
\setlength{\tabcolsep}{4pt}
\begin{tabular}{@{}lcccl@{}}
\toprule
Criterion cell & s3 & s4 & s5 & Frozen bar \\
\midrule
Amplitude z & 15.15 & 13.20 & 13.77 & $\ge3$ \\
MI excess (nats) & +0.1275 & +0.1025 & +0.1202 & $\ge0.10$ \\
$\zd$ vs emergent s1 ($\dexcess$) & +7.87 (+0.0806) & +5.92 (+0.0554) & +7.29 (+0.0734) & $\ge3$ \\
$\zd$ vs blocks s1 ($\dexcess$) & +7.26 (+0.0704) & +5.02 (+0.0452) & +6.64 (+0.0631) & $\ge3$ \\
\bottomrule
\end{tabular}
\end{table}

Replication control arms measured in the same collection: emergent s1 z=7.58 / excess +0.0471; blocks s1 z=10.88 / excess +0.0573. \textbf{Robustness disclosure (non-criterion):} against a second, stronger blocks control (s2; z=12.05, excess +0.0810 --- the unsupervised null strengthens further at this third arm), $\zd$ = +4.53 / \textbf{+2.21} (p=0.0145) / +3.86; the s4 cell lands below 3 and is reported as such. Per the frozen anti-rescue clause this robustness member does not enter, substitute for, or alter the gate verdict. We therefore state: at both scales, type supervision produces slot$\times$type organization that is statistically attributable to the supervision signal, above what unsupervised routing priors provide.

\subsection{The routing/readout boundary (H-$\alpha$): what the library is not}
\label{sec:boundary}

Three findings scope every editability word in this paper: (i) applying all 150/150 structural flips changes the answer readout by $\le2\times10^{-4}$ with exactly zero collateral --- the structural codes are \emph{not on the readout pathway}; (ii) ROME-style row fine-tuning is plastic at the edited point (0.855) but does not generalize to held-out variants (0.124); (iii) the tgC generalization probe fails its bar at both scales (0.124 at 22.6M; 0.217, three-seed mean, at 125M; frozen bar 0.30). We therefore describe the library as a \textbf{typed routing index} with exact state-level edit semantics, and we explicitly do \emph{not} claim behavioral editability of the world model.

This decoupling is not an architectural triviality. The library is the model's only writable causal state and the backbone consumes library-mediated representations: the architecture gave the explicit structure every condition to participate in readout, and training decoupled it anyway. We regard this boundary --- measured, replicated, and stable across a 5.6$\times$ scale window --- as a contribution: it maps precisely where an explicit-state architecture's guarantees end, and it falsifies the ``circuit-board'' intuition that exposed structure is necessarily on the computation's main path.

\begin{figure}[t]
\centering
\includegraphics[width=\textwidth]{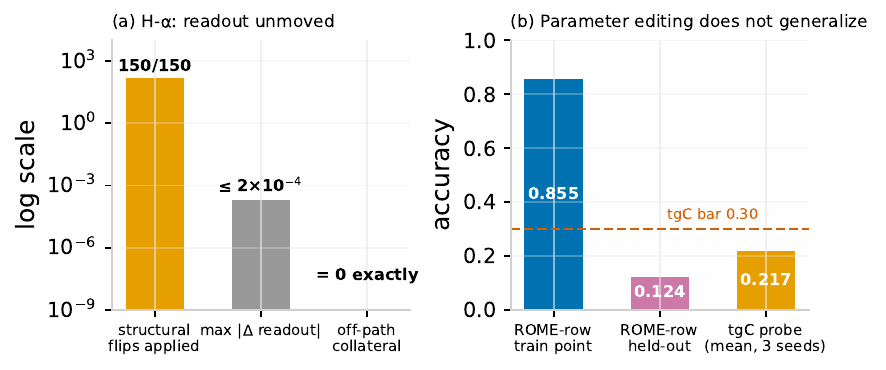}
\caption{Negative results, kept prominent: the library is editable \emph{as state}, not established as behaviorally editable. (a) 150/150 structural edits land on the library while the readout moves $\le2\times10^{-4}$ and collateral is exactly zero; (b) parameter-route editing is plastic at the train point (0.855) but does not generalize (0.124), and the tgC probe misses its 0.30 bar (0.217).}
\label{fig:boundary}
\end{figure}

\subsection{Locality at the state level (criterion S2)}
\label{sec:locality}

Paired flip edits (150 queries/seed, three seeds) at 125M: answer-readout change $\dy \le 3.4\times10^{-6}$ (criterion $\le10^{-3}$, met with $\ge2.5$ orders of magnitude margin) and \textbf{off-path collateral exactly 0.0 on all three seeds}. Edits to the library are exactly local at the state level, and the F6 pathway boundary --- structural slot codes do not feed the answer readout --- replicates at 5.6$\times$ scale without exception. The bounded parameter operator remains plastic at the training point (gain 0.884/0.935/0.917) but does not teach the readout to consume structural codes: tgC mean 0.217 $<$ 0.30 (capacity/generalization branch, preregistered tree). This is the project's central negative result and it is \emph{scale-invariant}; \S\ref{sec:boundary} states what it scopes.

\subsection{Zero LM cost (criterion S3)}
\label{sec:cost}

LM-quality gap vs the parameter-matched monolith: 0.0058 / 0.0082 / 0.0071 nats across the three type seeds (criterion $\le0.02$), and the 50-item Simpson-paradox behavioral battery passes 50/50 on both arms (criterion $\ge95\%$). The typed library costs nothing measurable in base capability.

\begin{figure}[t]
\centering
\includegraphics[width=\textwidth]{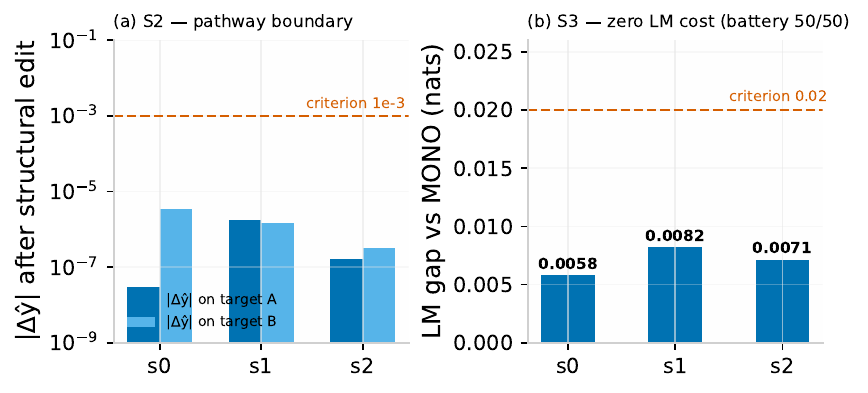}
\caption{Locality and cost. (a) Readout change is $\ge2.5$ orders of magnitude below the bar; collateral is not ``small'' but exactly zero. (b) LM gap within preregistered tolerance on all seeds.}
\label{fig:locality}
\end{figure}

\subsection{Bit-exact reversibility (criterion A22)}
\label{sec:revert}

The library state is bit-exactly revertible \textbf{as state}: the preregistered R1--R5 audit passes on all three seeds with zero failures --- 250/250 single-edit restores; 1,000/1,000 depth-20 stacked reverts; 1,250/1,250 backbone zero-touch comparisons; zero-contamination on every failed edit (147--150 sampled instances per seed; the count varies because an edge constructed to be absent exists in a few worlds, making the edit legal --- full counts in Table~\ref{tab:c4}, audit figure in Appendix~\ref{app:data}). Two properties matter more than the log: rollback completeness holds \textbf{by construction} (the audit buffers live in the model's \texttt{state\_dict}, so no snapshot can forget revertible state), and the toolchain itself passed a preregistered engineering-validation pass (A22.1; two defects caught and fixed) before any criterion data existed. Per our frozen wording this is a statement about toolchain state, not about behavioral edit semantics (\S\ref{sec:boundary}); the operator set is specified in \S\ref{sec:editinterface} and the full audit table in Appendix~\ref{app:data}.

\section{Analysis}

\subsection{Stability and seed variance}
\label{sec:stability}

Across permutation seeds, every arm's z is stable (worst sd 0.20, type s1). Raw type-arm excess spans 0.134--0.164 (between-seed sd $\approx$0.015): real seed-level variance exists, and three seeds is a small sample --- which is why the replication doubles the confirmatory base.

\subsection{A descriptive pooled view (no inferential weight)}

\begin{figure}[t]
\centering
\includegraphics[width=0.86\textwidth]{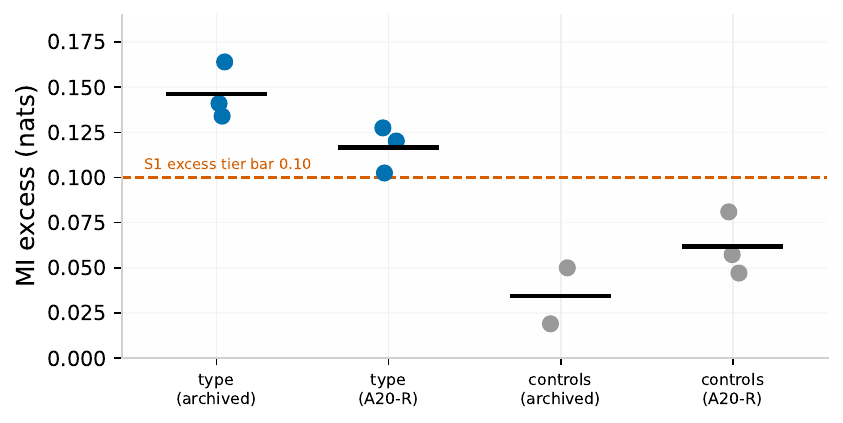}
\caption{Descriptive pooled view across collections (no confirmatory pooling). Type-arm excess replicates at a level far above every unsupervised control in both collections; the replication controls themselves sit above the archived ones --- the moving null of \S\ref{sec:organization}, visible in raw form.}
\label{fig:pooled}
\end{figure}

Figure~\ref{fig:pooled} pools, \textbf{descriptively}, the archived and replication collections: the six type seeds' excess values overlap across collections (archived 0.134--0.164; replication 0.1025--0.1275), while every unsupervised control sits far below (0.019--0.081). Per the preregistered rule there is \textbf{no confirmatory pooling} --- the archived and replication verdicts stand on their own frozen criteria, and this plot carries no inferential weight.

\section{Related work}

\textbf{Model editing.} Locate-and-edit methods modify knowledge inside distributed weights: ROME-style rank-one updates identify a decisive MLP layer and write a new key--value association \citep{meng2022locating}, and MEMIT-style successors extend this to thousands of simultaneous edits \citep{meng2023massediting}; MEND learns a hypernetwork that maps gradients to weight edits \citep{mitchell2022fast}. A large follow-up literature measures locality, generalization, and persistence trade-offs on behavioral benchmarks (CounterFact, zsRE, and derivatives: \citealp{meng2022locating,levy2017zeroshot}). Notably, localization itself has been shown not to be causal for editing outcomes \citep{hase2024does} --- a finding consonant with our routing/readout boundary, which offers one testable hypothesis for why such mismatches arise (\S\ref{sec:discussion}). Our work differs in \emph{where the editable object lives}: not an approximate subspace inside weights, but an explicit, enumerable state with construction-guaranteed locality (off-path collateral exactly zero) and bit-exact rollback. We deliberately do not frame this as a behavioral shoot-out --- those methods report behavioral generalization on their benchmarks, while our own H-$\alpha$ boundary shows our structural edits are not on the readout pathway at all. The correct contrast is the explicitness and auditability of the editable state, and the strength of the guarantees one can state before, rather than estimate after, the edit.

\textbf{Mechanistic interpretability and probing.} A parallel tradition discovers internal structure post-hoc: linear and nonlinear probes \citep{adi2017finegrained,hewitt2019designing}, sparse dictionary decompositions of superposed features \citep{elhage2022toy,bricken2023towards}, and circuit-level analyses of routing and attention \citep{olah2020zoom,wang2023interpretability}. These methods read structure out of models that were never asked to have any; our approach is complementary --- structure is built in, and the claim that it materialized is itself subjected to preregistered, permutation-tested measurement against unsupervised nulls, with pipeline-validation gates guarding the measurement instrument. Our moving-null finding (\S\ref{sec:organization}) is directly relevant to this literature: any organization or specialization claim that calibrates its null at one scale and evaluates at another risks inheriting a null that no longer applies.

\textbf{World models, memory, and structured state.} External-memory architectures \citep{graves2014neural,graves2016hybrid}, object-centric slot representations \citep{locatello2020objectcentric}, and algorithmic-state lines of work all endow models with structured, addressable state; causal-benchmark work supplies worlds with known ground-truth structure \citep{jin2023cladder,jin2024corr2cause,kiciman2023causal}. Our synthetic causal worlds follow the second tradition as a \emph{measurement instrument}: because ground-truth causal structure is known by construction, criteria of the form ``exactly zero collateral'' or ``bit-exact restore'' become checkable in a way that naturalistic benchmarks cannot support. Typed evidence routing as an inductive bias --- and the evidence-type supervision that organizes it --- connects both to recent work on evidence-type competition \citep{xun2026evidence} and to routing analyses in the mixture-of-experts literature \citep{shazeer2017outrageously,fedus2022switch,zhou2022mixture}.

\textbf{Preregistration and audit culture.} Our audit trail is methodologically kin to registered reports and preregistration practice in the experimental sciences \citep{nosek2018preregistration,chambers2013registered}, adapted to ML measurement \citep{forde2019scientific}: criteria archived before data, frozen decision trees, hash-chained scripts, and full publication of failed gates. Figure~\ref{fig:tree} shows the tree as executed; Appendix~\ref{app:cases} reports the two measurement case studies that shaped it.

\section{Discussion}
\label{sec:discussion}

\textbf{Scope, set by evidence.} Our contribution is not a SOTA improvement; it is the falsification of a widely held implicit assumption, plus the measurement machinery that made the falsification unavoidable. What the evidence supports: an explicit typed library can be trained at zero LM cost; it organizes under type supervision (at both scales, attributable to the supervision signal per \S\ref{sec:organization}); edits to it are exactly local and bit-exactly reversible at the state level; and its behavioral boundary is sharp, measured, and stable across a 5.6$\times$ scale window. What it does not support: behavioral editability (H-$\alpha$), and any claim that the unsupervised null is scale-free (\S\ref{sec:organization}).

\textbf{``If the library does not change behavior, what is it for?''} The result redirects rather than deflates model editing: the resistance to behavior change is not in the slot contents --- those are exactly editable --- but at the routing/readout connection. Editing research aimed at explicit state should target bridging H-$\alpha$ (teaching readouts to consume structural codes) rather than refining the state representation further; editing research aimed at distributed weights should treat our boundary as a baseline expectation, not an anomaly, given how readily the optimizer produced it here.

\textbf{Why does such decoupling exist? A speculation on residual-stream architecture.} We close with a hypothesis, explicitly labeled as such --- our data establish the boundary, not its cause. Discrete slot codes are high-precision, low-bandwidth objects, well suited as \emph{addresses}; coupling them directly to the readout would tie a sparsely-updated, discrete pathway to every output position, plausibly injecting gradient noise into the dense computation. A transformer trained by gradient descent may therefore find it favorable to keep discrete structure on the routing side while distributed weights carry the answer --- a separation of addressing from computation that acts as an implicit regularizer, discovered by the optimizer rather than imposed by us. This account is consistent with independent reports that localization need not determine editing outcomes \citep{hase2024does}, and it is falsifiable: interventions that bridge the boundary (e.g., training the readout to consume slot codes) should, under this account, trade off against training stability. Whether the same boundary arises for \emph{emergent}, non-engineered structure is the open question our instrument cannot answer --- and the one we consider most important.

\textbf{Limitations.} Synthetic causal worlds; two scales within one architecture family; three-seed arm budgets (six after replication); the 22.6M organization effect is moderate-grade (three-seed mean excess 0.0998 against the frozen strong-tier bar of 0.10 --- a 0.0002 miss recorded per the frozen tiering, not relabeled); the baseline-emergence confound (\S\ref{sec:organization}) is reported, not resolved; persistence under continued training (A21) remains untested --- the criterion is frozen and was unfrozen for execution on 2026-08-06 (archive A20.12/A20.21), but no A21 data exist yet.

\textbf{Why we report the pipeline event.} The binary-check failure at 125M, the suspended criterion, and the frozen-threshold handling are not embarrassments to minimize --- they are the audit system working in public. We submit that organization claims \emph{should} be made this way, and we release the machinery (scripts, hashes, decision trees) so others can hold us --- and themselves --- to it.

\section*{AI Use Statement}
A large-language-model assistant was used to help draft and copy-edit the manuscript, to adversarially proof-check its claims against the underlying experimental data and audit archive, and to develop and debug the training and evaluation code. All scientific claims, experimental design, data, analyses, and conclusions were made and verified by the author, who takes full responsibility for the content of this paper.

\section*{Reproducibility Statement}

Every quantitative claim in this paper is tied to a preregistered, machine-checkable criterion archived --- with md5-chained scripts --- before the data it governs; the full audit trail (entries A12--A22.3, including the failed gate and the execution incidents) is released as supplementary material. The causal-world generator, the frozen training recipe (Table~\ref{tab:arms}), all archived seeds (world, permutation, and training), and the evaluation code are released to enable exact reproduction: every world is reconstructible from its seed, and every reported number is traceable to a named archive entry. The A20-R replication protocol, including its power analysis and anti-rescue clauses, is included in full in the archive.

\section*{Ethics Statement}

This work uses only synthetic causal-world data; no human subjects, personal data, or scraped corpora are involved. The research goal --- auditable, exactly revertible state management for causal knowledge --- is oriented toward safer and more transparent model maintenance. We foresee no application-specific negative impacts beyond those of general-purpose language-model research; the benchmark's synthetic nature means conclusions about naturalistic settings require the further work we flag in Limitations.

\begingroup
\emergencystretch=1.5em
\raggedright
\bibliography{references}
\endgroup
\bibliographystyle{iclr2027_conference}

\appendix
\setlength{\textfloatsep}{8pt plus 2pt minus 2pt}
\setlength{\floatsep}{8pt plus 2pt minus 2pt}
\section{Preregistration archive (summary)}
\label{app:archive}

A-series protocol chain (each entry archived before the data it governs; md5-chained scripts; full text in repository):
\begin{itemize}
\item \textbf{A12}: pipeline-integrity incident --- a leaky aggregation fabricated significance; remediation: raw-pair permutation MI (\texttt{mi\_perm\_test}) + binary pipeline check. \emph{Lesson embedded in all later criteria.}
\item \textbf{A18/A19}: editing probes; H-$\alpha$ boundary established (strong form); contributions re-scoped (``editable'' removed; library = typed routing index); tgC branch decision.
\item \textbf{A20 (S1/S2/S3 at 125M)} $\to$ S2, S3 pass; pipeline gate fails (blocks z=2.38) $\to$ \textbf{A20.8--A20.14} event chain: diagnostics D-a--D-d; paired-attribution revision (frozen before data); 100-round adversarial review; A20-R powered replication protocol (frozen, locked); execution incidents (interrupted runs, env failure, deadlock) logged with zero data impact.
\item \textbf{A20-R (powered attribution replication)} $\to$ executed as frozen (T0--T6 unified chain; QC gates passed: 6/6 checkpoint-args assertions, bit-exact redundant collection); \textbf{gate passed on all nine criterion cells (2026-08-06, A20.21)} $\to$ \S\ref{sec:organization} BRANCH-A selected mechanically; robustness member (blocks s2) disclosed per the anti-rescue clause.
\item \textbf{A21 (persistence)}: drafted, reviewed 20 rounds, frozen; unfrozen 2026-08-06 upon the A20-R gate pass (A20.12 condition met); execution to be scheduled --- no A21 data exist yet.
\item \textbf{A22 (revert toolchain)}: R1--R5 criteria + engineering validation (2 defects caught and fixed before criterion data); unfrozen early by preregistered orthogonality ruling; \textbf{passed} (\S\ref{sec:revert}).
\end{itemize}

\section{Measurement case studies}
\label{app:cases}

\textbf{Case A12 (fabricated significance).} A leaky aggregation step once fabricated significance on null data; the remediation (raw-pair permutation MI plus a binary pipeline check) now guards every MI claim in this paper.

\begin{figure}[!ht]
\centering
\includegraphics[width=0.92\textwidth]{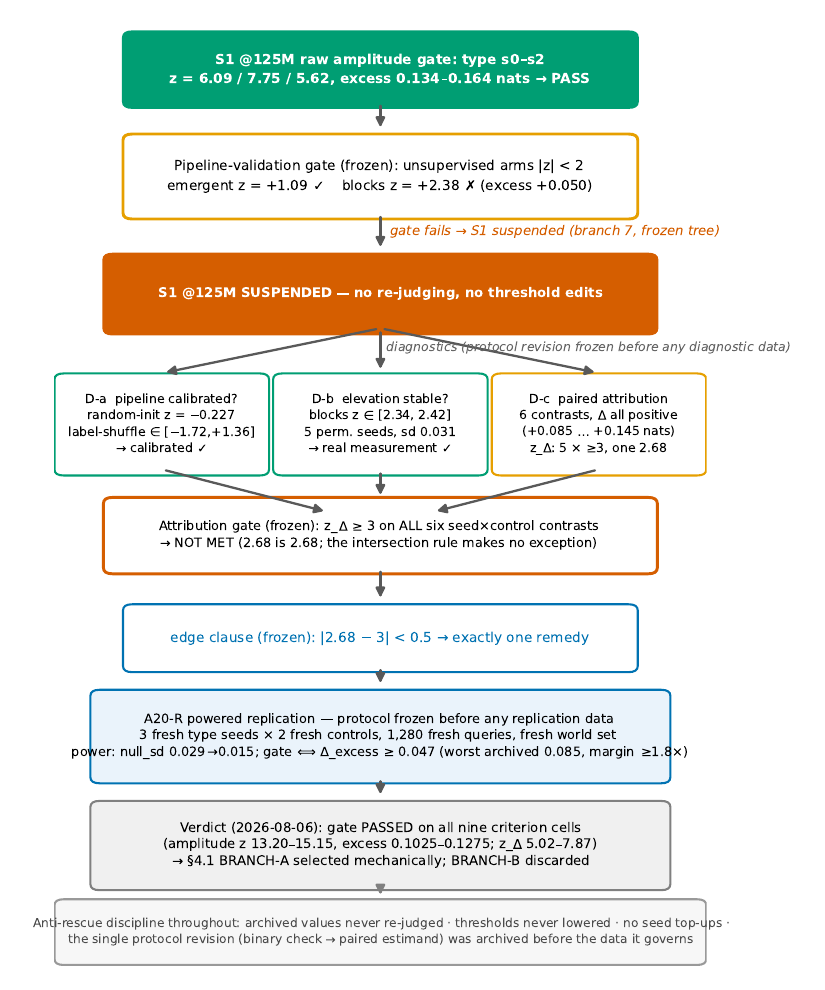}
\caption{The preregistered decision tree, as executed (Case A20). Every node is an archived event with archived values; every threshold shown was frozen before the data it governs.}
\label{fig:tree}
\end{figure}

\textbf{Case A20 (the moving null).} The 125M pipeline-gate failure; D-a calibration; D-b stability; paired-contrast revision; frozen-threshold discipline; powered replication design (null\_sd $\approx0.029$@320q $\to$ $\approx0.015$@1280q; gate $\Longleftrightarrow$ $\dexcess \ge 0.047$; observed worst case 0.085, margin $\ge1.8\times$). Figure~\ref{fig:tree} shows the tree as executed; Table~\ref{tab:attribution} is the full D-c table; both narrative branches were pre-frozen.

\section{Full data tables}
\label{app:data}
\setcounter{table}{0}\renewcommand{\thetable}{C\arabic{table}}

\begin{table}[H]
\caption{S1 raw measurements at 125M (archived, frozen; 320 queries, 2,000 permutations per arm).}
\label{tab:c1}
\centering\small
\begin{tabular}{@{}lccl@{}}
\toprule
Arm & z & MI excess (nats) & p \\
\midrule
type s0 & +6.09 & +0.141 & $<1/2000$ \\
type s1 & +7.75 & +0.164 & $<1/2000$ \\
type s2 & +5.62 & +0.134 & $<1/2000$ \\
emergent s0 & +1.09 & +0.019 & 0.1465 \\
blocks s0 & +2.38 & +0.050 & 0.012 \\
\bottomrule
\end{tabular}
\end{table}

\begin{table}[H]
\caption{D-a pipeline calibration (two independent null families, all inside $|z|<2$).}
\label{tab:c2}
\centering\small
\begin{tabular}{@{}lccl@{}}
\toprule
Null & z & excess & p \\
\midrule
random-init checkpoint & $-0.227$ & $-0.0015$ & 0.5345 \\
blocks s0 label-shuffle K=1 & +1.355 & +0.0284 & 0.0990 \\
blocks s0 label-shuffle K=2 & $-0.151$ & $-0.0032$ & 0.5525 \\
blocks s0 label-shuffle K=3 & $-1.720$ & $-0.0365$ & 0.9655 \\
blocks s0 label-shuffle K=4 & +0.205 & +0.0044 & 0.4030 \\
blocks s0 label-shuffle K=5 & $-0.924$ & $-0.0196$ & 0.8225 \\
\bottomrule
\end{tabular}
\end{table}

\begin{table}[H]
\caption{D-b permutation-seed stability (archived seed + four fresh seeds per arm).}
\label{tab:c3}
\centering\small
\begin{tabular}{@{}lllll@{}}
\toprule
Arm & archived z & four fresh-seed z & sd & range \\
\midrule
type s0 & 6.09 & +6.06 / +6.13 / +6.00 / +6.03 & 0.048 & [6.00, 6.13] \\
type s1 & 7.75 & +7.25 / +7.79 / +7.52 / +7.65 & 0.200 & [7.25, 7.79] \\
type s2 & 5.62 & +5.56 / +5.30 / +5.55 / +5.53 & 0.105 & [5.30, 5.62] \\
emergent s0 & 1.09 & +1.13 / +1.13 / +1.05 / +1.11 & 0.032 & [1.05, 1.13] \\
blocks s0 & 2.38 & +2.41 / +2.42 / +2.34 / +2.39 & 0.031 & [2.34, 2.42] \\
\bottomrule
\end{tabular}
\end{table}

\begin{table}[H]
\caption{A22 revert-toolchain audit counts (three seeds; all verdicts PASS, zero failures).}
\label{tab:c4}
\centering\small
\begin{tabular}{@{}lccc@{}}
\toprule
Check & s0 & s1 & s2 \\
\midrule
R1 single-edit bit-exact (50 worlds $\times$ 5 operators) & 250/250 & 250/250 & 250/250 \\
R2 depth-20 stacked rollback & 1000/1000 & 1000/1000 & 1000/1000 \\
R3 backbone zero-touch comparisons & 1250/1250 & 1250/1250 & 1250/1250 \\
R4 function-level deviation $\le10^{-6}$ & 250/250 & 250/250 & 250/250 \\
R5a empty-stack no-op contract & 50/50 & 50/50 & 50/50 \\
R5b failed-edit zero-contamination & 147/147 & 150/150 & 150/150 \\
\bottomrule
\end{tabular}
\end{table}

R5b's s0 count of 147 (vs 150) is constructive sampling, not a criterion gap: in three worlds the edge constructed to be absent actually exists, making the edit legal and hence not a failure-semantics instance. S2/S3 per-seed values appear in \S\ref{sec:locality}--\S\ref{sec:cost}; the A20-R replication table is Table~\ref{tab:a20r} (\S\ref{sec:organization}).

\begin{figure}[!ht]
\centering
\includegraphics[width=0.92\textwidth]{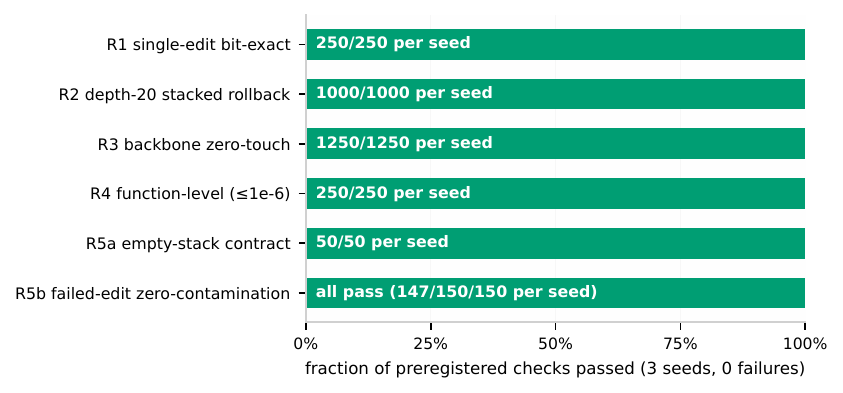}
\caption{Revert-toolchain audit results. Every preregistered check passes on every seed.}
\label{fig:revert}
\end{figure}

\section{Claims matrix}
\label{app:claims}
\renewcommand{\thetable}{\arabic{table}}\setcounter{table}{4}

\begin{table}[H]
\caption{Claims matrix (every row traceable to a named archive entry).}
\label{tab:claims}
\centering\footnotesize
\setlength{\tabcolsep}{3pt}
\begin{tabular}{@{}L{3.2cm}L{3.7cm}L{5.0cm}@{}}
\toprule
Claim & Criterion & Status \\
\midrule
Typed organization, 22.6M & S1 (per-seed $z\ge3$; frozen mean-excess tiers) & \textbf{Supported --- moderate grade} (z=4.47/5.99/7.02; excess 0.072--0.124, mean 0.0998) \\
Typed organization, 125M, attributed & S1 + paired-attribution gate & \textbf{Supported} (A20-R gate passed on all nine criterion cells, 2026-08-06; archive A20.21) \\
Exact edit locality & S2 (collateral $<0.01$) & \textbf{Supported} (0.0 exactly, $\times3$) \\
F6 pathway boundary scale-invariant & S2 ($\dy\le10^{-3}$) & \textbf{Supported} ($\le3.4\times10^{-6}$, $\times3$) \\
Zero LM cost & S3 ($\mathrm{gap}\le0.02 \wedge \mathrm{battery}\ge95\%$) & \textbf{Supported} ($\le0.0082$; 50/50) \\
Toolchain bit-exact rollback & A22 R1--R5 & \textbf{Supported} (all pass, $\times3$) \\
Unsupervised null moves with scale & pipeline gate + D-b/D-d diagnostics & \textbf{Observed} (blocks +0.050 nats at 125M vs $\approx0$ at 22.6M; cause confounded, reported as such) \\
Effect-size growth with scale ($\approx1.5\times$) & --- (not preregistered) & \textbf{Post-hoc observation}, labeled as such \\
Behavioral editability & --- & \textbf{Not claimed} (H-$\alpha$) \\
Persistence under continued training & A21 C1--C3 & [PENDING: criterion frozen; unfrozen 2026-08-06; execution to be scheduled] \\
\bottomrule
\end{tabular}
\end{table}

\end{document}